\documentclass[conference]{IEEEtran}
\IEEEoverridecommandlockouts
\usepackage{cite}
\def\BibTeX{{\rm B\kern-.05em{\sc i\kern-.025em b}\kern-.08em
    T\kern-.1667em\lower.7ex\hbox{E}\kern-.125emX}}
\usepackage{amsmath}
\usepackage[utf8]{inputenc}
\usepackage{bm}
\usepackage{amsfonts}
\usepackage[stable]{footmisc}
\usepackage{graphicx}
\usepackage{graphics}
\usepackage{amssymb}
\usepackage{algorithmicx}
\usepackage{algorithm}
\usepackage{algpseudocode}
\usepackage{newtxtext,newtxmath}        
\usepackage{multirow}
\usepackage{url}
\usepackage{lscape}
\usepackage{xcolor}

\newcommand{\R}{{\mathbb R}}

\newcommand{\N}{{\mathbb N}}

\newcommand{\be}{\begin{equation}}
\newcommand{\ee}{\end{equation}}
\newcommand{\ba}{\begin{array}}
\newcommand{\ea}{\end{array}}
\newcommand{\baa}{\left[\begin{array}}
\newcommand{\eaa}{\end{array}\right]}

\newcommand{\beqa}{\begin{eqnarray}}
\newcommand{\eeqa}{\end{eqnarray}}
\newcommand{\bt}{\begin{tabular}}
\newcommand{\et}{\end{tabular}}

\newcommand{\bi}{\begin{itemize}}
\newcommand{\ei}{\end{itemize}}
\newcommand{\bc}{\begin{center}}
\newcommand{\ec}{\end{center}}

\newtheorem{result}{Proposition}

\newtheorem{remark}{Remark}

\newtheorem{corollary}{Corollary}

\newcommand{\norm}[1]{\left\lVert#1\right\rVert}

\newcommand{\eref}[1]{(\ref{#1})}
\begin{document}

\title{Efficient and Parallel Separable Dictionary Learning
\thanks{
C. Rusu was supported by a grant of
the Ministry of Research, Innovation and Digitization, CNCS/CCCDI-UEFISCDI,
project number PN-III-P1-1.1-TE-2019-1843, within PNCDI III.
P. Irofti was supported by two grants of
the Ministry of Research, Innovation and Digitization, CNCS/CCCDI-UEFISCDI,
project number PN-III-P1-1.1-PD-2019-0825 and
project number PN-III-P2-2.1-PED-2019-3248, within PNCDI III.
}
}

\author{\IEEEauthorblockN{1\textsuperscript{st} Cristian Rusu}
\IEEEauthorblockA{\textit{Department of Automatic Control and Computers} \\
\textit{Faculty of Automatic Control and Computers},\\
\textit{University Politehnica Bucharest}\\
Bucharest, Romania}
\IEEEauthorblockA{\textit{Research Center for Logic, Optimization and Security (LOS)},\\
\textit{Department of Computer Science} \\
\textit{Faculty of Mathematics and Computer Science},\\
\textit{University of Bucharest}\\
Bucharest, Romania \\
\texttt{cristian.rusu@upb.ro}}
\and
\IEEEauthorblockN{2\textsuperscript{nd} Paul Irofti}
\IEEEauthorblockA{\textit{Research Center for Logic, Optimization and Security (LOS)},\\
\textit{Department of Computer Science} \\
\textit{Faculty of Mathematics and Computer Science},\\
\textit{University of Bucharest}\\
Bucharest, Romania \\
\texttt{paul@irofti.net}, 0000-0002-7541-4334}
}

\maketitle

\begin{abstract}
Separable, or Kronecker product, dictionaries provide natural decompositions for 2D signals, such as images. In this paper, we describe a highly parallelizable algorithm that learns such dictionaries which reaches sparse representations competitive with the previous state of the art dictionary learning algorithms from the literature but at a lower computational cost. We highlight the performance of the proposed method to sparsely represent image and hyperspectral data, and for image denoising.
\end{abstract}

\section{Introduction}

Dictionary learning (DL)~\cite{DL_book} aims at
finding a suitable overcomplete basis, or dictionary,
that best represents a given dataset: training signals 
vectorized as columns in $\bm{Y}$.
Based on this training matrix
we find the dictionary $\bm{D}$ that produces representations $\bm{X}$
such that $\bm{Y} \approx \bm{DX}$.
We mandate sparsity in the representations~\cite{Elad10}
which implies that a sample $\bm{y}$,
uses only a few columns, also called atoms, from $\bm{D}$
encoded in the representations. 
Popular algorithms include
Orthogonal Matching Pursuit (OMP)~\cite{PRK93omp}
for sparse representations
and
Approximate K-SVD (AK-SVD)~\cite{RZE08} or more recent adaptive stable techniques \cite{SEGHOUANE2018300}
for dictionary learning.

Often, samples have an intrinsic structure that is lost or weakened 
with standard DL approaches.
For example, in images and videos,
pixels and voxels have strong
vicinity based connections leading to certain patterns exploited
by standard signal processing tools.
Also, network or graph generated signals
suffer from the same loss of structure and recent DL work~\cite{Yankelevsky16_dualgraph,BPI20_ifac} has shown
that recovering the graph structure through the learning process can
significantly improve results.
Thus,
existing algorithms were adapted
to separable variants that maintain the data structure
by training separate dictionaries for each dimension~\cite{Hawe13_sedil,DCL17_sukro,Soltani2016Dec}
but also exploit the structure in order to gain better and faster
approximations~\cite{FWH12_2domp,ID19_pair}.

Previous work introduced parallel/distributed DL algorithms but these addressed non-separable problems~\cite{ID14_gpuksvd,7219480}
and introduced significant computational overhead~\cite{Dai12_simco}.
We continue this pursuit and propose a new parallel algorithm that provides a fast and distributed (with little communication overhead) separable DL solution based on the Tensor MOD (TMOD) approach~\cite{Fu14_jmdl}.

In the separable dictionary setting,
for a bidimensional~(2D) signal $\bm{Y}$
we now want to represent it using two dictionaries
$\bm{Y}\approx\bm{D}_1\bm{X}\bm{D}_2^T$.
The approximation quality is identical
$\| \bm{Y} - \bm{D}_1 \bm{X} \bm{D}_2^T \|_F =
\| \text{vec}\bm{Y} - (\bm{D}_2 \otimes \bm{D}_1)\text{vec}\bm{X} \|_2$,
if the one-dimensional~(1D) dictionary is
set as $\bm{D}=\bm{D}_2 \otimes \bm{D}_1$.
Here 
$\text{vec}\bm{M}$ denotes the column-order vectorization of matrix $\bm{M}$, $\norm{.}_F$ the Frobenius norm and $\otimes$ is the Kronecker product.
Throughout the paper we treat the 2D case,
but the results shown can be easily generalized to multiple dimensions \cite{7839993}.

In the context of DL,
separable or pairwise dictionaries are used to represent 2D signals
such as images.
The data is now stored as a set of $N$
samples $\bm{Y}_k \in \R^{m \times m}$, $k=1:N$.
Small dictionaries 
$\bm{D}_1 \in \R^{m \times n_1}$ (the left dictionary)
and
$\bm{D}_2 \in \R^{m \times n_2}$ (the right dictionary)
cater to the 2D data
such that they produce
the corresponding
$\bm{X}_k \in \R^{n_1 \times n_2}$ sparse representations
that are $s \in \N^*$ sparse
(the matrix $\bm{X}_k$ has at most $s$ nonzero elements).
We are now ready to formulate the pairwise DL optimization problem as
\be
\begin{aligned}
& \underset{\bm{D}_1,\ \bm{D}_2,\ \bm{X}}{\text{minimize}} & & \sum_{k=1}^N \|\bm{Y}_k - \bm{D}_1 \bm{X}_k \bm{D}_2^T\|_F^2 \\
  & \text{subject to} & &  \|\bm{X}_k\|_{0} \leq s,\ k = 1:N,\\
  & & & \| \bm{d}_{1i} \|_2 = 1\text{ and } \| \bm{d}_{2j} \|_2 = 1, \\
\end{aligned}
\label{pair_DL}
\ee
where $i=1:n_1$, $j=1:n_2$, $\bm{d}_{1i}$ and $\bm{d}_{2j}$
denote the normalized atoms of the two dictionaries.
Existing algorithms~\cite{FWH12_2domp,ID19_pair}
exploit the separable formulation
such that $\bm{D}_2 \otimes \bm{D}_1$ is never explicitly computed.

\section{Separable Dictionary Learning}


The earliest work on separable dictionaries, either in the context of dictionary learning or compressed sensing for images, was introduced in \cite{Skretting} and \cite{4801726}. Recently, there has been revived interested in the separable dictionary learning problem, both theoretically and from an algorithmic perspective.

SeDiL~\cite{Hawe13_sedil} was among the first to attack the separable DL problem by employing a highly complex gradient descent algorithm on smooth Riemannian submanifolds.
Keeping the data structure intact, thus multi-dimensional, has lead to many tensor-based algorithms~\cite{DCL17_sukro,HLP14_2dns,ZA15_ktsvd,Fu14_jmdl}
that employ various CANDECOMP/PARAFAC (CP) decomposition tactics to  update pairs of atoms or even whole dictionaries at once.
We note that, with the exception of \cite{Fu14_jmdl}, \cite{HLP14_2dns}, \cite{ZA15_ktsvd} and \cite{8892653},
tensor methods choose not to exploit the separable structure directly. The STARK algorithm~\cite{8313164} implicitly enforces Kronecker structure on the dictionary by solving a regularized convex relaxation of the hard rank-1 tensor recovery problem while TeFDiL~\cite{8892653} is a factorization based approach which imposes explicitly the Kronecker structure and finds the small dictionaries $\bm{D}_i$.

MOD~\cite{EAH99mod} solves the DL problem by viewing $\bm{Y}=\bm{DX}$ as a least squares problem~(LS) where the variable is $\bm{X}$ in the representation stage, and $\bm{D}$ in the dictionary update stage.
TMOD is the $n$-dimensional adaptation of the MOD~\cite{Skretting,doi:10.1002/widm.1108,6854345} algorithm.
Given
tensor $\bm{T}\in\R^{I_1 \times \dots \times I_n}$
and
matrix $\bm{M}\in\R^{J\times I_k}$,
let $\bm{T}_{(k)}$ be the mode-$k$ matricization
and let $\otimes_k$ be the mode-$k$ product
such that
$(\bm{T} \otimes_k \bm{M})_{(k)} = \bm{MT}_{(k)}$.
TMOD writes the $n$-dimensional DL problem as
$\bm{Y}=\bm{X}\otimes_1\bm{D}_1\otimes_2\bm{D}_2\otimes_3 \dots \otimes_n \bm{D}_n$,
where $\bm{Y}$ is the tensor containing $N$ samples of $m_1\times \dots \times m_n$ dimensions each
and $\bm{D}_i\in\R^{m_i \times n_i}$ is the dictionary associated to dimension $i$.
The right-hand variables are updated one at a time by solving a large LS problem. Similarly, the K-SVD algorithm has also been extended to the tensor setting~\cite{6854345}. As convolutional dictionaries are able to capture local structure in image data, separable convolutional dictionary  learning was also introduced recently \cite{8461884,8517082}.

In the sparse representation stage,
these algorithms usually employ the 2D-OMP algorithm~\cite{FWH12_2domp}.
The complexity reduction brought by the separable version
and its equivalence to 1D OMP~\cite{PRK93omp} is thoroughly demonstrated in \cite{ID19_pair} and \cite{FWH12_2domp}.

In coordinate descent fashion,
TKSVD~\cite{Fu14_jmdl} simultaneously updates atom pairs $i$, $j$ while keeping the other atoms fixed which can lead to an increase in the overall error \eref{pair_DL} as shown in \cite{ID19_pair}.
PairAK-SVD~\cite{ID19_pair}
is a direct non-tensor AK-SVD adaptation to the separable scenario 
which modifies the residual to update only one atom at a time.

STARK~\cite{8313164}, SuKro~\cite{DCL17_sukro}, and \cite{8892653} also extended 
TMOD
by writing $\bm{D}$ as a sum of Kronecker products.
Thus,
in \eref{pair_DL} we would write $\bm{D}  = \sum_{r=1}^R \bm{D}^{(r)}_2 \otimes \bm{D}^{(r)}_1$.
Dictionary update is also residual based 
but it involves a rank-$R$ problem solved via ADMM~\cite{Boyd11_admm}.
Earlier results in \cite{HLP14_2dns} are similar to SuKro and also propose a non-separable version based on CP decomposition. The latest in this line of work is a DL algorithm that learns sums of $R$ Kronecker products of $K = 2$ terms at a time~\cite{10.1007/978-3-319-93764-9_42}. Finally, the work in \cite{8892653} described an online approach to the separable dictionaries learning problem.

While we focused on algorithmic developments,
we also mention theoretical efforts made to understand the behavior and limits of these procedures
like
the local identifiability of the Kronecker-structured dictionaries~\cite{8892653,8361034,8849698}
and
sample complexity analysis of these dictionaries~\cite{7541479,7953008,ShakeriSBbook} that was shown to be significantly lower than for unstructured dictionaries~\cite{7378975}.
Recently, matrix factorization were used to provide guarantees for global optimality~\cite{doi:10.1137/18M121976X}.

\section{The proposed algorithm}

In this section, we describe the proposed solution to the separable dictionary learning problem. We aim to solve the problem in \eqref{pair_DL} in three steps: 1) with $\bm{D}_1$ and $\bm{D}_2$ fixed we update all $\bm{X}_k$ using the 2D-OMP algorithm (alternatively, other sparse recovery algorithms can be used in this step but we choose an OMP approach due to its advantageous numerical properties); 2) for $\bm{D}_1$ and all $\bm{X}_k$ fixed we compute the optimal $\bm{D}_2$, the minimizer of the objective function in \eqref{pair_DL}; and 3) analogous to the previous step, we find the optimal $\bm{D}_1$. We choose 2D-OMP over 1D-OMP in order to keep memory usage low and avoid vectorization operations that need to take place during the 1D sparse recovery algorithms. In this paper, we do not focus on developing new 2D sparse approximation methods but our contribution lies in developing new, efficient, ways of finding the two dictionaries $\bm{D}_1$ and $\bm{D}_2$. Our dictionary update rules are based on the following result and corollary.

\begin{result}[Optimal $\bm{D}_2$ update]
Let $\bm{D}_1$ be fixed and denote $\bm{Z}_k = \bm{D}_1 \bm{X}_k$ with all $\bm{X}_k$s fixed, then the minimizer of the objective function in \eqref{pair_DL} is
\begin{equation}
    (\bm{D}_2^\star)^T = \bm{W}_2 (\sum_{k=1}^N \bm{Z}_k^T \bm{Z}_k)^{-1} (\sum_{k=1}^N \bm{Z}_k^T \bm{Y}_k).
\end{equation}
\noindent{\textit{Proof.}} Ignore for now the norm constraints and the objective function of \eqref{pair_DL} develops into
\begin{equation}
    \begin{aligned}
        & \sum_{k=1}^N \norm{ \bm{Z}_k \bm{D}_2^T - \bm{Y}_k  }_F^2 
        =  \\
    & \norm{ (\sum_{k=1}^N \bm{Z}_k^T
    \bm{Z}_k)^{1/2} \bm{D}_2^T - (\sum_{k=1}^N \bm{Z}_k^T \bm{Z}_k)^{-1/2}(\sum_{k=1}^N \bm{Z}_k^T \bm{Y}_k) }_F^2 + C.
    \end{aligned}
\end{equation}
Here $C = \text{tr}( \sum_{k=1}^N \bm{Y}_k^T \bm{Y}_k  - (\sum_{k=1}^N \bm{Y}_k^T \bm{Z}_k)(\sum_{k=1}^N \bm{Z}_k^T \bm{Z}_k)^{-1}$ $(\sum_{k=1}^N \bm{Z}_k^T \bm{Y}_k) )$ does not depend on $\bm{D}_2$. Since the minimization above reduces to a standard least squares problem, the minimizer (assuming $\sum_{k=1}^N \bm{Z}_k^T \bm{Z}_k$ has full rank, which happens almost always as $N \gg m$) is
    $(\bm{D}_2^\star)^T = (\sum_{k=1}^N \bm{Z}_k^T \bm{Z}_k)^{-1} (\sum_{k=1}^N \bm{Z}_k^T \bm{Y}_k)$. Let $\bm{W}_2$ be a $n_2 \times n_2$ diagonal matrix such that $\bm{D}_2^\star\bm{W}_2$ has normalized columns, then the optimal updates are $\bm{D}_2^\star \leftarrow \bm{D}_2^\star \bm{W}_2$ and $\bm{X}_k \leftarrow \bm{X}_k \bm{W}_2^{-1}$, without affecting the sparsity pattern (the objective function is minimized with the LS solution and the normalization diagonal cancels in the products $\bm{X}_k (\bm{D}_2^\star)^T$) $.\hfill \blacksquare$
\end{result}

\begin{corollary}[Optimal $\bm{D}_1$ update]
Analogously to Result 1, denoting and fixing $\bm{T}_k = \bm{X}_k \bm{D}_2^T$ we have that the minimizer of the objective function in \eqref{pair_DL} is given by
    $\bm{D}_1^\star =  (\sum_{k=1}^N \bm{Y}_k \bm{T}_k^T) (\sum_{k=1}^N \bm{T}_k \bm{T}_k^T)^{-1} \bm{W}_1,$ where $\bm{W}_1$ is a diagonal matrix of size $n_1 \times n_1$ chosen such that $\bm{D}_1^\star$ has normalized columns and update $\bm{X}_k \leftarrow \bm{W}_1^{-1}\bm{X}_k$ for all $k.\hfill \blacksquare$
\end{corollary}

\begin{algorithm}[t]
		\caption{Distributed Separable Dictionary Learning}
		\begin{algorithmic}
\State\hspace{-0.4cm} \noindent \textbf{Require:} $\bm{Y}\in\R^{m\times m \times N}$,
		             $\bm{D}_{1,2}\in\R^{m \times n_{1,2}}$,
		             and $s,p,K \in \N^*$
		             
\State\hspace{-0.4cm} \noindent \textbf{Result:} $\bm{D}_1, \bm{D}_2$, and $\bm{X}\in\R^{n_1 \times n_2 \times N}$ (distributed at the nodes)

\State\hspace{-0.4cm} \noindent \textbf{Setup:} Split as evenly as possible the $N$ data points $\{ \bm{Y}_i \}_{i=1}^p$ and distribute initial $\bm{D}_{1,2}$ among the $p$ processing nodes

\State\hspace{-0.4cm} \noindent \textbf{Main loop, for $1,\dots,K$:} 
		    
			\State \textbf{1. } Each node $i$ computes the $s$-sparse representations $\bm{X}_k$ for all its data points $\bm{Y}_k$ and its partial sums $\bm{P}_i = \sum_k \bm{T}_k \bm{T}_k^T$ and $\bm{R}_i = \sum_k \bm{T}_k \bm{Y}_k^T$;
			
			\State \textbf{2. } Master node accumulates $\bm{P}_i$ and $\bm{R}_i$ from all nodes $i$, sums $\bm{P} = \sum_{i=1}^p \bm{P}_i$ and $\bm{R} = \sum_{i=1}^p \bm{R}_i$, and then computes $\bm{D}_1 = \bm{P}^{-1} \bm{R} \bm{W}_1$, according to Corollary 1. Master node distributes $\bm{D}_1$ to all $p$ nodes;
			
			\State \textbf{3. } With the freshly received $\bm{D}_1$, each node $i$ computes the $s$-sparse representations $\bm{X}_k$ for all its data points $\bm{Y}_k$ and its partial sums $\bm{M}_i = \sum_k \bm{Z}_k^T \bm{Z}_k$ and $\bm{N}_i = \sum_k \bm{Z}_k^T \bm{Y}_k$;
			
			\State \textbf{4. } Master node accumulates $\bm{M}_i$ and $\bm{N}_i$ from all nodes $i$, sums $\bm{M} = \sum_{i=1}^p \bm{M}_i$ and $\bm{N} = \sum_{i=1}^p \bm{N}_i$, and then computes $\bm{D}_2 = \bm{M}^{-1} \bm{N} \bm{W}_2$, according to Result 1. Master node distributes $\bm{D}_2$ to all $p$ nodes.
		\end{algorithmic}
		\label{alg:dl}
\end{algorithm}

\begin{figure*}[t]
	\centering
	\includegraphics[trim=510 340 670 100, clip, width=0.6\textwidth]{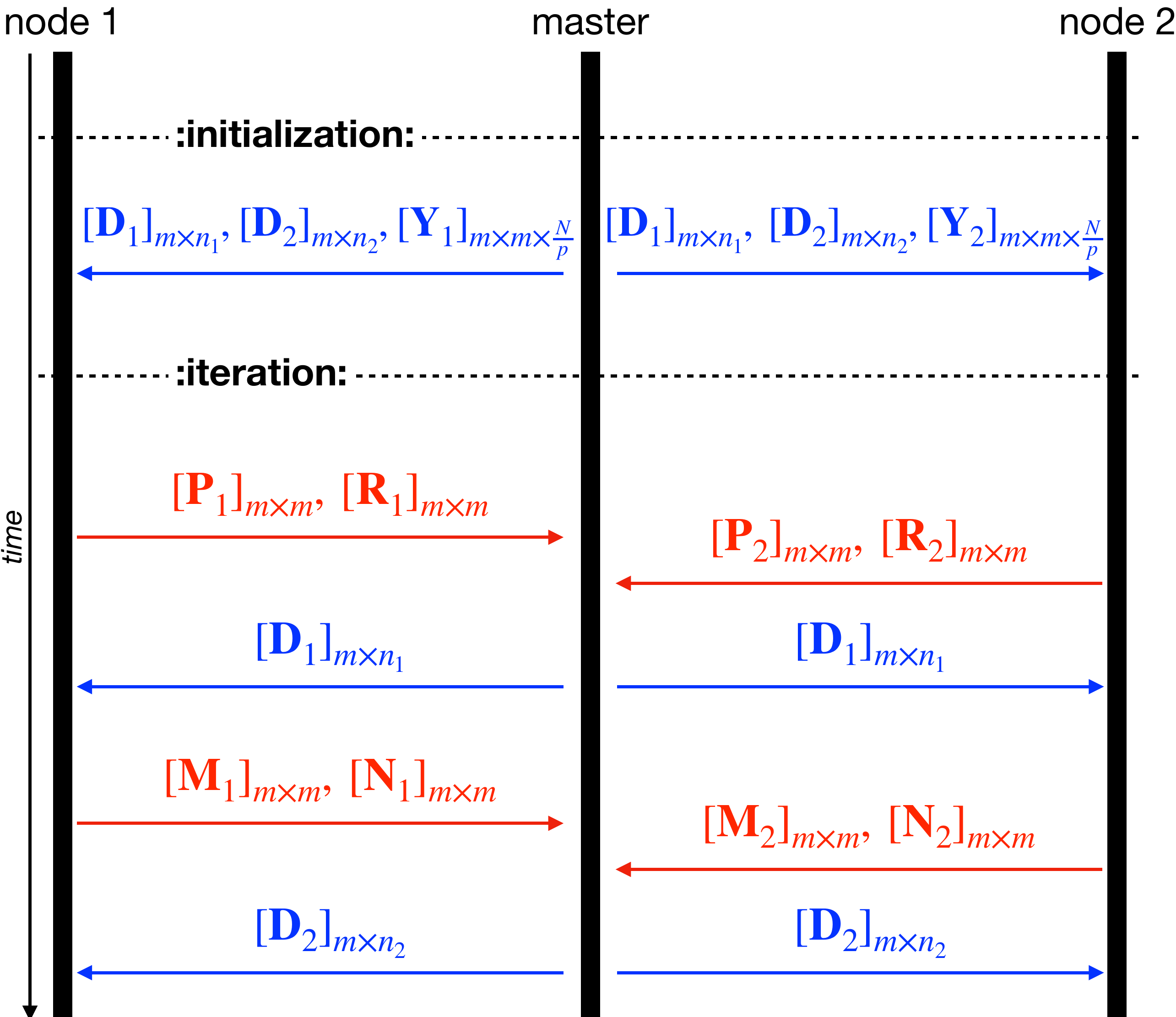}
	\caption{Assuming the overall dataset is split in $p$ parts, we show the thread communication diagram with a master and two of the processing nodes highlighting the distributive nature of the proposed algorithm. We explicitly show all the dimensions of the matrices being communicated.}
	\label{fig:figure0}
\end{figure*}

Denoting $\bm{G}_1 = \bm{D}_1^T \bm{D}_1$
and
$\bm{G}_2 = \bm{D}_2 \bm{D}_2^T$,
we note that we will compute 
$\bm{Z}_k^T \bm{Z}_k = \bm{X}_k^T \bm{G}_1 \bm{X}_k$
and
$\bm{T}_k \bm{T}_k^T = \bm{X}_k \bm{G}_2 \bm{X}_k^T$.
We describe the full proposed distributed alternating optimization method in Algorithm~\ref{alg:dl}.
All dictionary updates guarantee a monotonic decrease in the objective function value
but the same is not true about 2D OMP, in general. Algorithm 1 is an explicit, efficient parallel/distributed implementation of the classic TMOD approach for separable dictionary learning.
In Figure \ref{fig:figure0}, we provide a thread communication diagram that shows the initialization process and one iteration of the proposed method. We describe a scenario with $p$ nodes but, for simplicity, show the communication/computation behavior only for two nodes highlighting also the dimensions of the matrices that are transferred -- which do not depend on the size of the dataset $N$. In general, dictionary learning algorithms can be trivially parallelized in the sparse approximation step, i.e., the $N$ sparse solutions $\bm{X}_k$ are computed separately. The issue is that each $\mathbf{X}_k$ of size $m \times m$ needs to be transferred/copied to a central node for the dictionary update step. For our method, just four $m \times m$ matrices for each processing node are transferred and therefore the communication cost is lowered considerably, i.e., $O(m^2p)$ instead of $O(m^2N)$. As we will show experimentally, this allows the overall proposed algorithm to scale very well with the number of processing nodes $p$.

\begin{remark}[Learning orthonormal dictionaries]
    If we impose in \eqref{pair_DL} the additional constraint that we are learning orthonormal dictionaries, i.e., $n_1 = n_2 = m$, $\bm{D}_1^T \bm{D}_1 = \bm{D}_1 \bm{D}_1^T = \bm{I}_m$ and $\bm{D}_2^T \bm{D}_2 = \bm{D}_2 \bm{D}_2^T = \bm{I}_m$, then the least squares problem in Result 1 becomes an orthogonal Procrustes problem \cite{Proc} whose solution is given by a singular value decomposition (for both $\bm{D}_1$ and $\bm{D}_2$). We call Algorithm~\ref{alg:dl} - Orthonormal, the same approach as Algorithm~\ref{alg:dl} but with the Procrustes updates for the dictionaries. Orthonormal dictionaries also improve the numerical complexity of the algorithm. The update formulas for the dictionaries become: $\bm{D}_1^\star = \bm{U}_1\bm{V}_1^T$ where $(\sum_{k=1}^N \bm{Y}_k^T \bm{D}_2 \bm{X}_k)(\sum_{k=1}^N \bm{X}_k \bm{X}_k^T) =  \bm{U}_1 \bm{\Sigma}_1 \bm{V}_1^T$ is given by the singular value decomposition (SVD) and $\bm{D}_2^\star = \bm{U}_2\bm{V}_2^T$ where $(\sum_{k=1}^N \bm{Y}_k^T \bm{D}_1 \bm{X}_k) (\sum_{k=1}^N \bm{X}_k^T \bm{X}_k)  = \bm{U}_2 \bm{\Sigma}_2 \bm{V}_2^T$ again by the SVD. Furthermore, 2D-OMP is no longer necessary as the sparse representations are computed as $\bm{X}_k = \mathcal{T}_s(\bm{D}_1^T \bm{Y}_k \bm{D}_2)$, where $\mathcal{T}_s$ is an operator that keeps only the $s$ largest (in absolute value) entries of the input matrix.$\hfill \blacksquare$
\end{remark}
\begin{remark}[On the computational complexity of Algorithm~\ref{alg:dl}]
    We will highlight in the results section that the proposed method has a running time competitive against previously proposed algorithms from the literature. For simplicity let us assume that $n_1 = n_2 = m$. We focus on the dictionary updates for a single iteration of Algorithm~\ref{alg:dl} (in total we perform $K$ iterations). First, notice that computing $\bm{D}_1^\star$ and $\bm{D}_2^\star$ takes about $\frac{26}{3} m^3 + 12smN + 8m^2 N + 4m^3N$ operations: the first term includes the two Cholesky decompositions ($\frac{2}{3} m^3$), then the four back-substitutions ($4m^3$) used to solve the least squares problems for symmetric positive definite matrices with multiple right-hand sides and the two matrix-matrix multiplication to compute $\bm{G}_1$ and $\bm{G}_2$ ($4m^3$), the second term is the cost of building all $\bm{Z}_k^T \bm{Z}_k$ and $\bm{T}_k \bm{T}_k^T$ ($8smN$) and the cost of computing all the sparse and non-sparse matrix products ($4smN$ and $4m^3N$, respectively) and then summing them up ($8m^2N$) needed in Result 1 and Corollary 1. Because in general that $N \gg m$ we see that the computational complexity is dominated by the construction of the matrix product sums.
    
    From a computational perspective, tensor methods (TMOD, TKSVD, etc.) generally use a form of
    CP decomposition with alternating LS.
    Note that 
    computing the inverse in the vectorized case costs $O(m^6)$ operations,
    while in the 2D separable case it only takes $O(m^3)$.
    
    Furthermore, note that the proposed approach is highly parallelizable: computing the $N$ matrix products summations is distributed among multiple computing units (either CPUs or GPUs) with minimal cross-communication (only a partial summation matrix of size $m \times m$ needs to be communicated). The same holds for the calculations of representations $\bm{X}_k$ which can be done locally at each processing unit without the need of communicating them explicitly.$\hfill \blacksquare$
\end{remark}

\begin{remark}[Generalization to $n$ dimensions]
    Our result can be easily extended to more than two dimensions.
    We earlier described how TMOD generalizes MOD in $n$ dimensions.
    Our method also holds in $n$ dimensions using
    the property of the mode-$k$ product:
$\bm{Y}=\bm{X}\otimes_1\bm{D}_1\otimes_2\bm{D}_2\otimes_3 \dots \otimes_n \bm{D}_n \iff \bm{Y}_{(k)} = \bm{D}_k \bm{X}_{(k)}(\bm{D}_n \otimes \bm{D}_{n-1} \otimes \dots \otimes\bm{D}_1)^T$.
 Denoting with $\bm{T}$ the fixed dictionaries in the parenthesis we arrive at Corollary 1.
    $\hfill\blacksquare$
\end{remark}

\section{Simulation results \footnotemark}
\label{sec:sims}

\footnotetext{Matlab~\&~Python code at \url{https://github.com/pirofti/ParallelSeparableDL}}
In this section we showcase the results achieved by the proposed algorithms (orthonormal and general Kronecker dictionaries) and compare with the state of the art.

\begin{table}
	\caption{Average running times (seconds) for methods in Figure \ref{fig:figure1}.}
	\label{tab:runningtimes}
	\bc \bt{l c  c  c  c}
	\hline
\multirow{2}{*}{Method}& 
\multicolumn{2}{c}{$n = 8$} & \multicolumn{2}{c}{$n=16$} \\
& $s = 6$ & $s = 8$ & $s = 6$ & $s = 8$\\
\hline
	proposed, general $(p = 1)$ & 21 & 22 & 33 & 38 \\
    proposed, ortho $(p=1)$ & 18 & 20 & 18 & 22 \\
    PairAK-SVD & 140 & 171 & 205 & 266 \\
    SuKro & 23 & 27 & 47 & 56 \\
    TeFDiL & 8 & 10 & 19 & 24 \\
    AK-SVD & 10 & 14 & 16 & 21 \\ \hline
	\et \ec
\end{table}

\begin{figure*}[t]
	\centering
	\includegraphics[trim=120 1 70 15, clip, width=1.1\textwidth]{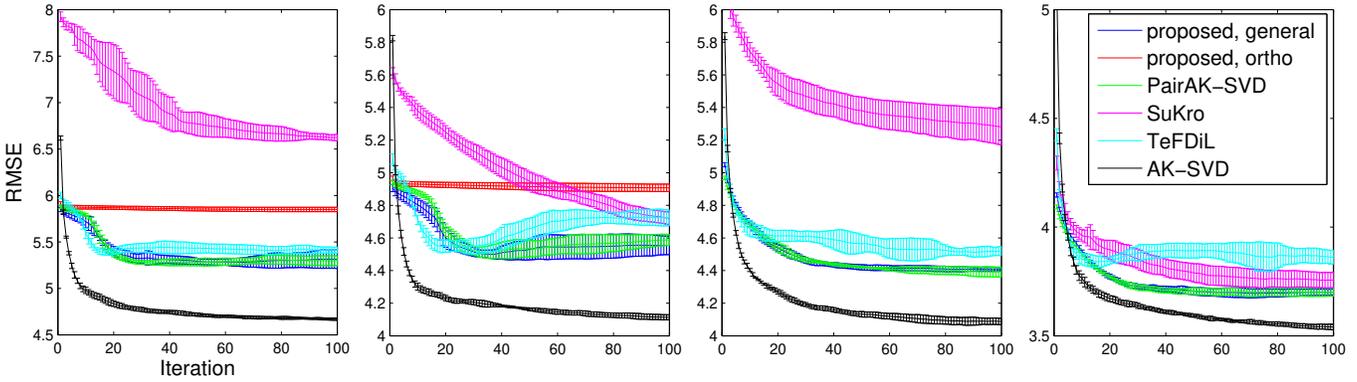}
	\caption{Mean and standard deviation RMSE over 100 iterations with 5 random realizations of datasets. From left to right: 1) each dataset consists of $N = 9216$ randomly chosen non-overlapped $8 \times 8$ image patches from a set of $12288$ patches, $n=8$ and $s=6$; 2) analogous to 1) for $s=8$; 3) analogous to 1) for $n=16$ and $s=6$; 4) analogous to 3) for $s=8$.}
	\label{fig:figure1}
\end{figure*}
\begin{figure*}[t]
	\centering
	\includegraphics[trim=120 1 70 15, clip, width=1.1\textwidth]{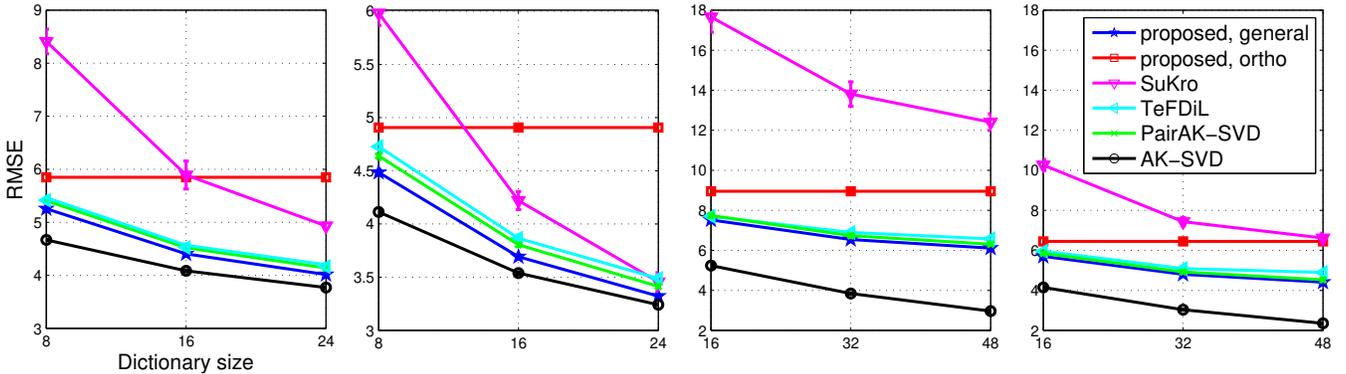}
	\caption{Mean and standard deviation RMSE against the size $n$ of the dictionaries over 5 random realizations of datasets. From left to right: 1) each dataset consists of $N = 9216$ randomly chosen non-overlapped $8 \times 8$ image patches from a set of $12288$ patches, $s=6$; 2) analogous to 1) for $s=8$; 3) each dataset consists of $N = 2304$ randomly chosen non-overlapped $16 \times 16$ image patches from a set of $3072$ patches, $s=8$; 4) analogous to 3) for $s=16$.}
	\label{fig:figure2}
\end{figure*}

First, in Figures \ref{fig:figure1} and \ref{fig:figure2} we show the RMSE achieved for various patches sizes $m$, dictionary sizes $n$, and dataset sizes $N$. Table \ref{tab:runningtimes} shows the running times of these methods
(proposed methods run on $p=1$ nodes) averaged over 5 rounds. As we are learning a relatively low number of dictionary parameters (roughly $O(mn)$) we choose $N$ on the order of $10^3$ and $10^4$ image patches. The proposed general approach equals the performance of PairAK-SVD, but at a fraction of the computational cost (on average $\times 7$ speedup as compared to PairAK-SVD and $\times 1.7$ against SuKro). TeFDiL is always faster than the general approach but the results are on average slightly worse. We would like to mention that the dictionary update step of TeFDiL involves large matrix-matrix multiplication (both matrices of size $m^2 \times N$) and solving linear systems of size $m^2 \times m^2$ with $N$ right-hand sides, both of which are highly parallelizable in Matlab (in fact, running Matlab with the flag -singleCompThread leads to a 33-50\% increase in the running time of TeFDiL). Therefore, one might say that TeFDiL is already ``half-paralellized'' (sparse representations are still computed sequentially). While TeFDiL is easy to parallelize (in a multi-thread fashion), the dictionary update step needs all the sparse representations $\bm{X}$ explicitly in the same processing node, so the communication cost in a distributed system would be high, as compared with our proposed approach where only small matrices need to be communicated (see Remark 2). The proposed orthogonal approach performs worse (of course due to the orthogonality constraints) but it is even faster -- note that we report running time for 100 iterations but the proposed orthogonal approach converges much faster, mostly before 20 iterations. For perspective, we show the error achieved by an unstructured dictionary via AK-SVD \cite{AEB06} which is faster than the proposed method in its serial ($p=1$) implementation. We now explore the speed-up benefits of the proposed method.

\begin{table}[t]
 \caption{Speedup achieved by the parallel implementation of the proposed algorithms ($m=n$). For ortho $s=m$, general $s=8$.}
\bc \bt{l c | c c c c c c}
\hline
\multirow{2}{*}{Method} & \multirow{2}{*}{$m$}
& \multicolumn{6}{c}{Number of processors ($p$)} \\
    
&& 1 & 2 & 4& 6& 8& 12\\
\hline

\multirow{3}{*}{\parbox{1.4cm}{proposed \\ ortho}} & 8 & 217 sec & $\times$2 & $\times$4.3& $\times$5.6& $\times$8& $\times$10.3\\
& 16 & 86 sec& $\times$2& $\times$3.9& $\times$5.4& $\times$7.8& $\times$10.8 \\
& 32 & 46 sec& $\times$2& $\times$3.8& $\times$5.1& $\times$7.1& $\times$10.6\\

\hline

\multirow{3}{*}{\parbox{1.4cm}{proposed \\ general}} & 8 & 440 sec & $\times$0.6 & $\times$1.1 & $\times$1.4 & $\times$2 & $\times$2.6 \\
& 16 & 450 sec& $\times$1.5 & $\times$3.1 & $\times$3.8 & $\times$5.1 & $\times$5.4  \\
& 32 & 469 sec& $\times$2 & $\times$3.4 & $\times$4 & $\times$4.6 & $\times$5 \\

\hline

\et \ec
\label{tab:speedup}
\end{table}

We have touted the efficient parallel implementation of Algorithm~\ref{alg:dl} and now in Table \ref{tab:speedup} we show the speedups achieved for both proposed approaches for various $m \times m$ patches. The datasets, for each $m \in \{8, 16, 32\}$, have $N = 128000$, $N = 32000$, and $N = 8000$  patches, respectively. For the proposed orthogonal case we scale almost precisely with the number of cores while for the proposed general case the speedup is less impressive. The latter observation is also due to the OMP implementation from the Python scikit-learn toolkit, which covers almost all the running time of the algorithm. We tested on an Intel(R) Xeon(R) CPU E5-2630 v4 @ 2.20GHz with 16 GB of RAM and 16 cores.

\begin{table*}[t]

\caption{Average denoising PSNR(dB) and SSIM for standard images over 5 realizations. Best results are in boldface (PSNR) or underlined (SSIM).}
\bc
\renewcommand{\arraystretch}{1.3}
\bt{c c c c c c c c c c c c}

\hline
\multirow{2}{*}{$\sigma_\text{noise}$ / PSNR} &
\multirow{2}{*}{Method}

  &\multicolumn{2}{c}{lena}
  &\multicolumn{2}{c}{barbara}
  &\multicolumn{2}{c}{boat}
  &\multicolumn{2}{c}{peppers}
  &\multicolumn{2}{c}{house} \\

  & &
  PSNR & SSIM &
  PSNR & SSIM &
  PSNR & SSIM &
  PSNR & SSIM &
  PSNR & SSIM \\
\hline

\hline\multirow{3}{*}{$ 5$ / $34.1557$ }
& $m=8, n=16$, general & 37.724& 0.9358& 33.045& 0.9376& 34.017& 0.9031& 36.372& 0.9059& 37.384& 0.9277 \\
& $m=n=8$, ortho & \textbf{38.381}& \underline{0.9407}& \textbf{37.895}& \underline{0.9611}& \textbf{36.977}& \underline{0.9352}& \textbf{37.354}& \underline{0.9194}& \textbf{38.904}& \underline{0.9459} \\
& PairAK-SVD~\cite{ID19_pair} & 37.912& 0.9367& 37.118& 0.9588& 35.987& 0.9210& 36.420& 0.9066& 38.038& 0.9380 \\
\hline\multirow{3}{*}{$10$ / $28.1322$ }
& $m=8, n=16$, general & 35.132& 0.9041& 31.126& 0.9044& 32.276& 0.8622& 34.367& 0.8705& 34.912& 0.8898 \\
& $m=n=8$, ortho & \textbf{35.237}& \underline{0.9048}& \textbf{33.949}& \underline{0.9280}& \textbf{33.334}& \underline{0.8738}& \textbf{34.553}& \underline{0.8730}& \textbf{35.316}& \underline{0.8944} \\
& PairAK-SVD~\cite{ID19_pair} & 35.185& 0.9046& 33.720& 0.9260& 33.183& 0.8724& 34.4076& 0.8708& 35.116& 0.8927 \\
\hline\multirow{3}{*}{$20$ / $22.1105$ }
& $m=8, n=16$, general & 31.988& 0.8557& 28.387& 0.8400& 29.723& 0.7858& 31.915& \underline{0.8323}& 32.275& 0.8532 \\
& $m=n=8$, ortho & 31.911& 0.8548& 29.833& 0.8599& 29.832& 0.7857& 31.836& 0.8313& 32.096& 0.8507 \\
& PairAK-SVD~\cite{ID19_pair} & \textbf{32.006}& \underline{0.8563}& \textbf{30.027}& \underline{0.8634}& \textbf{29.980}& \underline{0.7906}& \textbf{31.930}& \underline{0.8323}& \textbf{32.308}& \underline{0.8535} \\
\hline\multirow{3}{*}{$30$ / $18.5868$ }
& $m=8, n=16$, general & 29.877& 0.8174& 26.921& 0.7823& 27.886& 0.7252& 30.116& 0.8031& 30.159& 0.8211 \\
& $m=n=8$, ortho & 29.848& 0.8165& 27.396& 0.7909& 27.834& 0.7230& 30.055& 0.8021& 30.012& 0.8186 \\
& PairAK-SVD~\cite{ID19_pair} & \textbf{29.937}& \underline{0.8182}& \textbf{27.637}& \underline{0.7973}& \textbf{27.989}& \underline{0.7281}& 30.093& 0.8024& \textbf{30.250}& \underline{0.8220} \\
\hline\multirow{3}{*}{$50$ / $14.1505$ }
& $m=8, n=16$, general & 27.358& 0.7552& 24.211& 0.6695& 25.486& 0.6410& \textbf{27.570}& \underline{0.7531}& 27.296& 0.7609 \\
& $m=n=8$, ortho & 27.358& 0.7550& 24.449& 0.6780& 25.459& 0.6397& 27.513& 0.7520& 27.230& 0.7592 \\
& PairAK-SVD~\cite{ID19_pair} & \textbf{27.375}& \underline{0.7557}& \textbf{24.604}& \underline{0.6842}& \textbf{25.518}& \underline{0.6421}& 27.559& 0.7527& \textbf{27.302}& \underline{0.7614} \\

\hline

\et
\ec
\label{tab:denoising}
\end{table*}

In Table \ref{tab:denoising} we reproduced the denoising experiments described in \cite{ID19_pair} with the proposed methods:
we use $512\times 512$ images from the USC-SIPI database~\cite{sipi}
and denoise via error driven OMP with $\varepsilon = 1.15\sigma_{\text{noise}}\sqrt{m}$;
if the target error is not met, OMP is stopped when sparsity $s=m/2$ is reached.
The results are averaged over 5 realizations of noise.
We recorded negligible differences between realizations. This is standard practice in the literature~\cite{EladAharon06_denoising}.

For training we used $N=4000$ patches of $8\times 8$ pixels with which we learned $m=8$ by $n=16$ dictionaries with target sparsity $s=6$.
The chosen images and noise levels are often used in the literature~\cite{Hawe13_sedil,sipi,EladAharon06_denoising}.
Denoising
is performed on all $N = 255025$ overlapping patches
and
results are compared to the original image and measured in terms of peak signal-to-noise ratio (PSNR) and structural similarity index (SSIM)~\cite{WBSS04_ssim}.

\begin{table}
	\caption{Speedups achieved for Hyperspectral Experiments.}
	\label{tab:hyperspectral}
	\bc \bt{lcccccc}
	\hline
    & \multicolumn{6}{c}{Number of processors ($p$)} \\
	Method & 1 & 4 & 8 & 10 & 12 & 16\\ \hline
	ortho & 20532 sec & $\times 3.8$ & $\times 7.6$ & $\times 9.4$ & $\times 10.8$  & $\times 12.4$  \\
    general & 325030 sec & $\times 2.1$ & $\times 5.5$ & $\times 8$ & $\times 9.1$ & $\times 10.5$ \\ \hline
	\et \ec
\end{table}

Finally, in Table~\ref{tab:hyperspectral} we show DL execution times and speedups for the proposed methods
on the Indian Pine 220-layered $614\times 1848$ hyperspectral image from~\cite{hyperspectral_purr}.
The layers are divided among the $p$ CPUs (in total $N=3913140$ image patches) and the $m=8$ by $n=8$ dictionaries are trained for 100 iterations with $s=16$. Even in this large scale experiment we observe scaling behavior similar to the results in Table \ref{tab:speedup}.
\section{Conclusions}
In this paper we proposed two highly parallelizable and distributed algorithms for separable dictionary learning based on least squares dictionary updates. We show experimentally that the algorithms scale excellently with the number of cores or processing nodes and are competitive with the current state of the art separable dictionary learning methods for sparse representations. This approach opens the possibility of learning dictionaries given hundreds of thousands or millions of training signals in a reasonable amount of time. 


\bibliographystyle{IEEEbib}
\bibliography{bib}
\end{document}